\documentclass{article}
\usepackage{graphicx}
\graphicspath{ {./images/} }
\usepackage{float}
\usepackage{wrapfig}
\restylefloat{figure}

\usepackage[final]{corl_2020} 

\title{Planning Paths Through Unknown Space by Imagining What Lies Therein}

\author{
  Yutao Han, Jacopo Banfi, and Mark Campbell\\
  Department of Mechanical and Aerospace Engineering\\
  Cornell University\\ 
  United States\\
  \texttt{\{yh675, jb2639, mc288\}@cornell.edu} \\
}

\begin{document}
\maketitle


\begin{abstract}
    This paper presents a novel framework for planning paths in maps containing unknown spaces, such as from occlusions. Our approach takes as input a semantically-annotated point cloud, and leverages an image inpainting neural network to generate a reasonable model of unknown space as free or occupied. Our validation campaign shows that it is possible to greatly increase the performance of standard pathfinding algorithms which adopt the general optimistic assumption of treating unknown space as free.
\end{abstract}


\section{Introduction}

Planning paths for robots in environments with unknown spaces, such as from occlusions, is a challenging task due to the difficulty in modeling what actually lies therein. The customary approach to tackling this challenge in traditional grid-based representations is to assume any unknown cells as free. Any discrepancy between such an optimistic assumption and the actual map developed while executing the path and collecting information is used to trigger the computation of a new plan~\cite{koenig2002d}.

The very nature of this problem is online, making an approach based on constant replanning reasonable. We humans typically use scene context to reason about unknown spaces. To reduce running into dead ends, humans {\em predict} possible realizations of unseen space and plan paths accordingly, thereby reducing the number of replans. The goal of this work is to endow robots with a way of reasoning about unknown spaces, as humans do, in order to plan more efficient paths.

\begin{figure}[h!]
  \centering
  \includegraphics[scale=0.37]{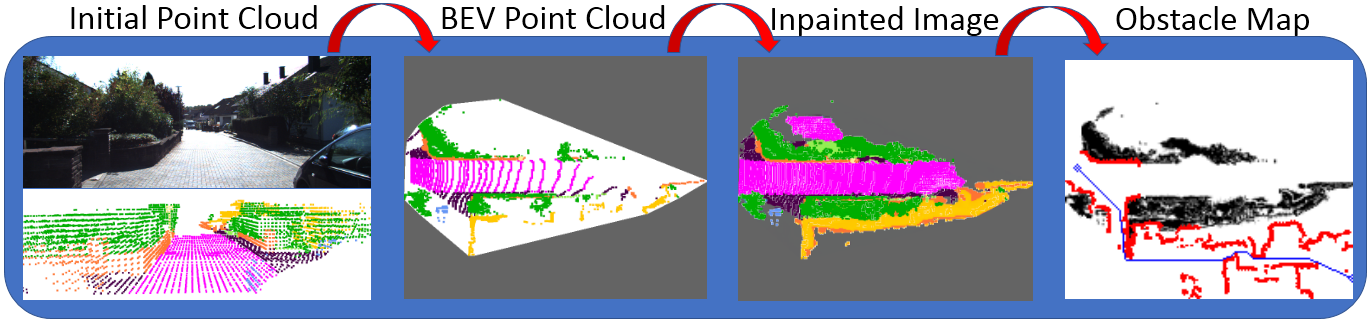}
  \caption{from left to right. Current partial map represented as a point cloud annotated with semantic labels (e.g. purple = road); bird's eye view of the annotated point cloud, where white corresponds to unlabeled pixels whose semantic class will be predicted via image inpainting; inpainted image; obstacle map associated with the inpainted image (white = free space, black = obstacles), shown along with obstacles discovered online (red) and final path to the goal (blue) in one of our simulations.}
  \label{fig:corl_beauty_image}
\end{figure}

Our approach, sketched in Figure~\ref{fig:corl_beauty_image}, takes as input a partial map in the form of a point cloud with semantic annotations, which can be obtained from a single lidar scan or a single stereo image pair. The semantic point cloud is converted into a bird's eye view (BEV) image, and the input into an image inpainting neural network~\cite{yu2018generative}. The network fills in empty pixels with semantic labels in order to obtain a reasonable prediction of an obstacle map for path planning.

The main contributions of this paper are: 

\begin{itemize}
    \item a novel framework for modeling unknown/occluded spaces for path planning using image inpainting to generate more informed maps
    \item the generation of a dataset for training an image inpainting network suitable for planning in outdoor urban environments.
\end{itemize}

The proposed framework is compared against a typical planner which treats all unknown space as free. Experimental results demonstrate the novel framework greatly reduces (1) the lengths of final paths which are traversed through occlusion heavy environments, (2) the number of replans online, and (3) the number of times the planner replans before it discovers no path is possible.


\section{Related Work}
Planning paths for mobile robots in the presence of a fully known, completely deterministic map is a well-understood problem in robotics. The classical approach leverages a grid-based representation of the environment, and uses a graph-search algorithm to plan a distance-optimal path to the goal. Among these, A*~\cite{hart1968formal} is the most widely used due to its efficiency. When the map is not fully known in advance, the customary expedient for adapting classical pathfinding algorithms is that of treating all the unknown cells as free, and computing new plans as needed. New algorithms were developed to handle the replanning phase more efficiently (D*~\cite{stentz1994} and its variants~\cite{stentz1995,koenig2002d}); however, unknown space has always been treated as free up to the present day (see, for example,~\cite{ryll2019efficient,tordesillas2019faster}). 

Learning-based approaches for navigation in partially unknown environments have started being explored only recently. Ref.~\cite{stein2018learning} focuses on partially mapped indoor environments represented as grids, and presents a method for selecting the ``frontier'' (i.e. the boundary between free and unknown space) to guide the robot in order to reach the goal. A utility function for the frontiers is learned via a neural network, but can only be used in indoor environments and with traditional (dense) partial grid maps. Ref.~\cite{dshppc2020} presents a next-best view planning approach that addresses unknown terrain semantics; however, the geometric map of the environment is assumed to be known in advance. This problem is also being actively investigated by the deep reinforcement learning community~\cite{mirowski2016learning,tai2017virtual,zhang2017deep,kahn2018self}. These works give up explicit map representations in favor of other features learned from experience. While these methods demonstrate promising initial results, they only work on simple indoor environments~\cite{tai2017virtual,kahn2018self} or mazes~\cite{mirowski2016learning,zhang2017deep}. 

In this work, we use an image inpainting network~\cite{yu2018generative} to fill in a bird's eye view (BEV) of an initial sensor measurement of an urban environment. The initial measurement has unobserved regions due to occlusions and sensor resolution, which make robust planning difficult. By filling in these regions with a data-driven inpainting network, a more informed map can be used for planning. While there are other works demonstrating the use of image inpainting to model urban environments ~\cite{Lu2020SemanticForeground,Purkait2019SeeingBehind,Schulter2018LearningTL}, these papers do not actually demonstrate their use in path planning tasks. In addition, these works inpaint behind segmented foregound object classes, which can be restrictive as occluded regions behind background classes can also be informative. For example, a building may be labeled as a background class, but it would be useful for a planner to reason about what is behind the building.

\section{Technical Approach}
Figure \ref{fig:corl_beauty_image} shows the flow chart of the proposed framework. Initially, a raw image is taken. Depth data is acquired either through stereo depth or from a lidar scan. Semantic segmentation~\cite{deeplabv3plus2018} is then performed on the image or lidar point cloud~\cite{behley2019iccv}. The depth and semantic labels are mapped to a point cloud which is rendered from a BEV for use as a map for a planner. Due to geometries in urban scenes, there will likely be unobserved pixels due to occlusions or sensor resolution. These unobserved pixels are then filled in using general adversarial network (GAN) based inpainting. The inpainted output of the network is then used as a more informed map for a planner.

\subsection{Image Inpainting}

Image inpainting is a technique traditionally used for restoring old or damaged pictures and paintings. The goal is to modify the images to be visually realistic to a viewer and restore the original state of the image ~\cite{Harrouss2019Inpaint, Bertalmio2000Inpaint}. Generally, the user labels regions of the image to be filled in, and the inpainting algorithm automatically fills in those regions. Image inpainting techniques include patch-based techniques, diffusion-based techniques, the use of convolutional neural networks (CNN), and the use of generative adversarial networks (GAN) ~\cite{yu2018generative, Harrouss2019Inpaint}. GAN-based inpainting networks simultaneously train a generative network to hallucinate images and a discriminative network to evaluate the hallucinated images. GAN-based methods are used in this paper due to their recent superior performance~\cite{yu2018generative, Yeh2017Inpaint}.

The GAN-based network used in this paper improves on previous work by introducing a contextual attention layer that utilizes distant image features during training to improve inpainting performance by reducing blurry textures and distortions~\cite{yu2018generative}. Typical inpainting applications include filling in missing parts of an outdoor scene or a persons face. In this work, the goal is to fill in a bird's eye view (BEV) of a semantic map such as the one pictured in Figure \ref{fig:corl_beauty_image} (middle left).

A set of $C$ semantic classes are defined for the input and output images. Each pixel is defined as $p_c(i,j)$, where $i$ and $j$ are the image coordinates of the pixel and $c$ is the semantic class of the pixel. In the input image the pixel can additionally be part of the unobserved class due to sensor range or occlusions. The goal of inpainting is to fill in the unobserved pixels (white) with a color corresponding to one of the $c$ semantic classes. The inpainted image is then used as a more informed map for the path planner.


To use the inpainted image as a map for planning, the set of $C$ semantic classes are divided into subsets of obstacle classes $C_o$ and non-obstacle classes $C_n$, where $C = C_o \cup C_n$ and $C_o \cap C_n = \emptyset$. Pixels belonging to $C_o$ are labeled as obstacles and pixels belonging to $C_n$ are considered free space.

\subsection{Path Planner}

Any planner suitable for traditional grid-based planning can be used in our framework; as the inpainting will improve the performance of each approach (for example, A*~\cite{hart1968formal}, Theta*~\cite{daniel2010theta}, or D*~\cite{koenig2002d}). A*~\cite{hart1968formal} is used in the experimental section of this paper given its extensive usage. The inpainted image is converted into a 2D obstacle map where each pixel is a node. Pixels belonging to $C_o$ (such as wall or vegetation) are labeled as obstacles (black), and pixels belonging to $C_n$ (such as sidewalk or road) are labeled as free space (white). As the robot executes the path, the occupied and free spaces are continuously recomputed (e.g. at a given frequency) based on the most recent sensory information (and hence obstacle map). Local planning incorporating the robot's kinematics (and possibly dynamics) can be easily performed by setting a local goal along the computed path, and is out of the scope of this work.



\subsection{Planning with Lidar Data}

In the proposed model, a dataset is required which includes pairs of data samples: an initial observation (Figure \ref{fig:SemKITTI} (left)), and a filled in ground truth semantic map (Figure \ref{fig:SemKITTI} (right)). The ground truth map is used for network training. Our framework (Figure \ref{fig:corl_beauty_image}) can be used with any lidar sensor that creates a semantically segmented point cloud~\cite{qi2016pointnet, behley2019iccv}. The key challenge is labeling ground truth filled in semantic maps for network training. Given adequate training data, the framework can easily be applied to any environment with unobserved space with any grid-based path planner.


\subsubsection{Lidar Urban Dataset Generation}
\label{sec:dataset}

\begin{wrapfigure}{H}{0.58\textwidth}
  \includegraphics[scale=0.35]{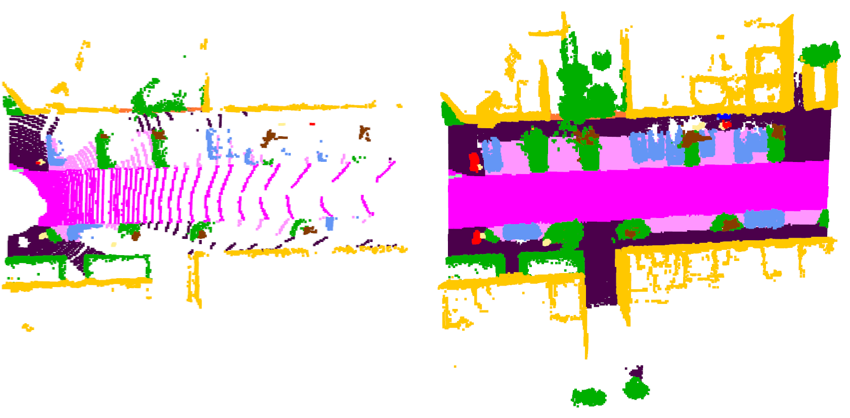}
  \caption{Left: The semantic input point cloud of an initial lidar scan. Right: The ground truth completed semantic map. \vspace{.1in}}
  \label{fig:SemKITTI}
\end{wrapfigure}

The recent SemanticKITTI dataset~\cite{behley2019iccv}, provides the tools to generate a dataset with pairs of data samples such as the one in Figure \ref{fig:SemKITTI}. More specifically, for each of the 22 sequences in the KITTI odometry dataset~\cite{geiger2012cvpr}, SemanticKITTI conglomerates all the lidar scans together into a single point cloud and manually labels 28 semantic classes. The point cloud is then voxelized and used for voxel-based semantic scene completion tasks. After every five time steps in each sequence, a ground truth semantic scene completion voxel map is provided within a predetermined volume of 256$\times$256$\times$32 voxels (with voxel resolution of 0.2m) in front of the car. The SemanticKITTI dataset also provides individual lidar scans within the same volume for each time step in each sequence along with the manually labeled semantics. The semantic lidar scans and completed voxel maps are shown in Figure \ref{fig:SemKITTI} with voxels converted to 3D points. These images are rendered from a BEV for the path planner applications. Semantic classes for moving objects and unidentifiable pixels are removed from the renderings as a static environment is assumed.

The goal of inpainting is then to fill the empty spaces in Figure \ref{fig:SemKITTI} (left) with the semantics in Figure \ref{fig:SemKITTI} (right); these unknown spaces are a function of occlusions or sensor resolution. To be compatible with semantic labels generated from a color camera image, only points which would be seen from a camera frame are kept. In order to do this with the SemanticKITTI dataset, all 3D points in the scan and completed map are projected into the camera frame:
\begin{equation}
x = Tr * P * X
\end{equation}
where $x$ is the coordinates of a point in the image frame, $Tr$ is the transformation from lidar coordinates to the rectified camera coordinates, $P$ is the projection matrix after image rectification, and $X$ represents the coordinates of a point in the lidar frame. All transformation matrices are provided by the KITTI dataset~\cite{geiger2012cvpr}. Only points which are projected into the image frame are kept. The points are then reprojected into lidar coordinates.

\begin{wrapfigure}{L}{0.7\textwidth}
  \includegraphics[scale=0.5]{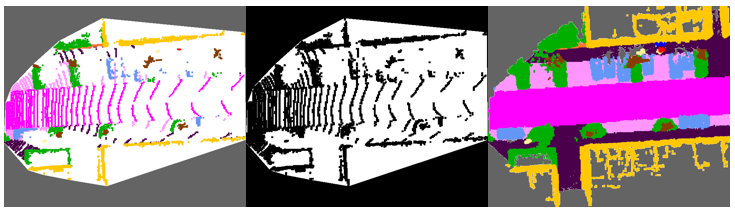}
  \caption{An example from the generated dataset of point clouds rendered from a BEV. Left: Input semantically labeled point cloud with occluded and unknown spaces in white. Middle: Mask of image to be inpainted. Known pixels are black and pixels to be filled are white. Note that grey (unobserved) pixels in the left image are labeled as known in the image mask as they are not the targets for inpainting. Right: Ground truth labeled semantic point cloud.}
  \label{fig:data_example}
\end{wrapfigure}

An example of a time step from the generated dataset is shown in Figure \ref{fig:data_example}. Because the GAN-based inpainting network uses all white pixels as pixels to be inpainted during training, when rendering the semantic point clouds from a BEV, all unknown pixels are rendered as grey. The adversarial network trains itself by blanking out sampled rectangles from the training image by setting the color in those rectangles as white. Therefore, in the original training image there should not be an excessive amount of white pixels as that would interfere with the network training. Figure \ref{fig:data_example} (left) shows the initial input image to be filled via inpainting. White pixels are required to be filled by the network. A convex hull is fit to the initial lidar scan and unknown pixels inside that convex hull are colored white. Only unknown pixels within the convex hull are designated for inpainting because the network is designed to inpaint unknown pixels using contextual information. In general, the network empirically performs poorly when a majority of pixels are designated for inpainting. Figure \ref{fig:data_example} (middle) shows the input mask to the network, where black pixels are known and white pixels are unknown. Figure \ref{fig:data_example} (right) is the ground truth filled in semantic map which is used for the actual training of the network.

\begin{figure}
  \centering
  \includegraphics[scale=0.36]{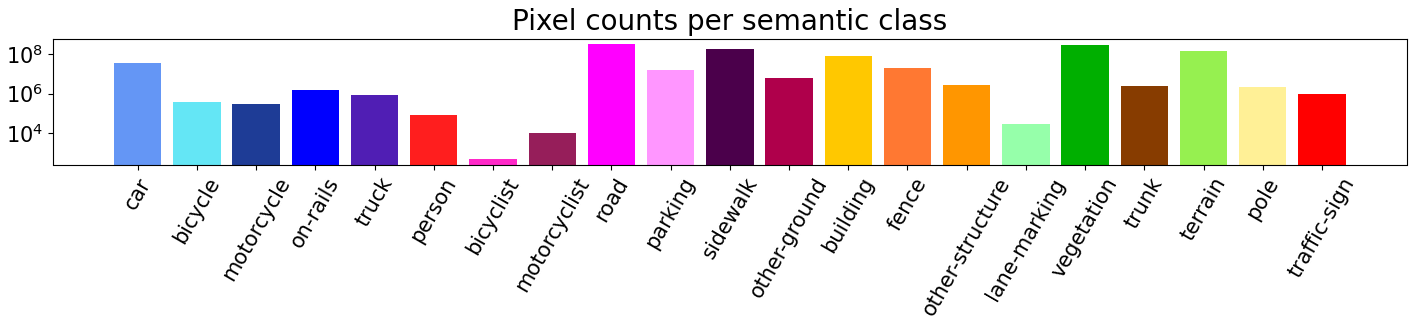}
  \caption{A distribution of the semantic pixel labels of the training dataset of the first ten sequences. Each semantic class is shown in the same color as displayed in the dataset. Note that the y-axis is log scale. \vspace{-.2in}}
  \label{fig:class_distribution}
\end{figure}

SemanticKITTI provides the first 11 of the 22 labeled KITTI odometry sequences publicly. In this work, the first ten sequences are used for training, and the eleventh sequence is used for testing. The fully processed dataset (all 11 sequences) of images shown in Figure \ref{fig:data_example} consists of 4,649 total pairs of training and input images. There are 241 images in the eleventh sequence which are used for testing. Two semantic classes of bus and other-vehicle did not have any corresponding labeled pixels. Figure \ref{fig:class_distribution} shows the distribution of semantic pixel labels in the training dataset of the first ten sequences. 

\subsection{Planning with Stereo Camera Data}
\label{sec:stereo_data}

The proposed novel framework can also use stereo camera data as input. Given the SemanticKITTI dataset, lidar depth and ground truth semantic labels are used for experimental purposes. However, the framework has the flexibility to also use stereo depth and semantic segmentations from any network. While the SemanticKITTI dataset contains ground truth for lidar depth which is used for the quantitative evaluation, a qualitative evaluation of semantic map inpainting with only a stereo camera as a sensor is performed to show the potential of the framework to be used with stereo depth.

The pipeline for using stereo depth is the same as for using lidar depth (Figure \ref{fig:corl_beauty_image}). Stereo results are expected to be worse than lidar due to quadratically increasing error in disparity with respect to depth. However, stereo cameras are frequently more convenient than lidar so it is important to have the flexibility to use stereo cameras in the framework.

For evaluation with stereo depth, learned stereo disparity from PSMNET~\cite{chang2018pyramid} is used due to its high performance on KITTI benchmarks. However, other stereo depth algorithms could be used without loss of generality. To perform semantic segmentation, a DeepLabv3+ network~\cite{deeplabv3plus2018} pretrained on Cityscapes~\cite{Cordts2016CVPR} is used. While the Cityscapes dataset is not necessarily the same as the KITTI dataset, there is a large overlap of classes and the SemanticKITTI dataset uses a pretrained network from Cityscapes for image segmentation.

\section{Experimental Evaluation}

Both qualitative and quantitative experimental evaluations are conducted for lidar depth input. Only qualitative evaluations are conducted for stereo depth input as there is no ground truth data for stereo depth. The output of the inpainting network is qualitatively evaluated based on its visual comparisons to the environment being reconstructed. Inpainting on input images generated from lidar scans and stereo cameras are evaluated. Quantitative evaluation is conducted on the inpainting accuracy compared with ground truth using the mean Intersection-over-Union (mIoU)~\cite{behley2019iccv}. The ability of the inpainting network to generate better informed maps for path planning is quantitatively evaluated by comparing the (1) number of planner steps taken, (2) the number of times the planner replans, and (3) the number of times the planner replans before discovering no path is possible.

\subsection{Network Training}

The GAN-based network is trained using output (filled in semantic maps) ground truth images generated of the first ten sequences from SemanticKITTI dataset. There are a total of 4,408 images in the training dataset (700$\times$580 pixels each). The network is trained on a GeForce GTX 1080 Ti GPU with 11 GB of memory. The network was trained for two and a half days.

\begin{figure}
  \centering
  \includegraphics[scale=0.75]{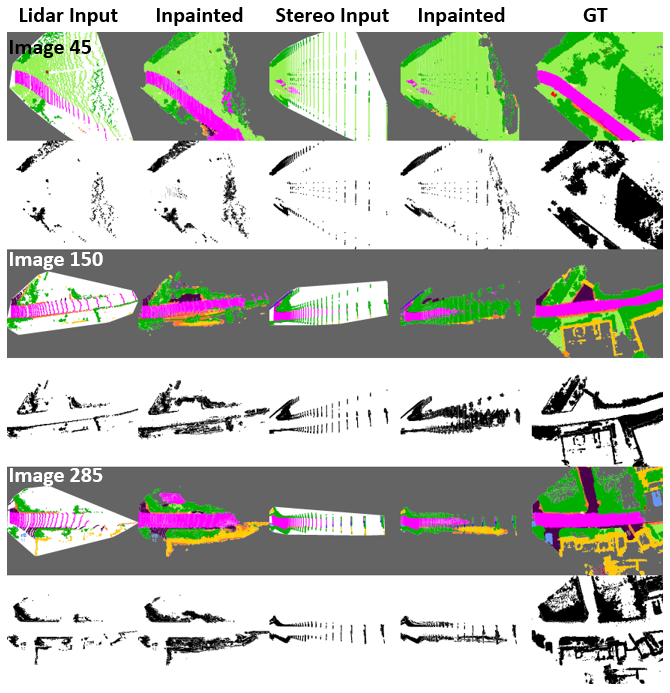}
  \caption{Qualitative evaluation of the inpainting network performance. Odd rows are the RGB semantic maps and even rows are the obstacle maps generated from the semantic BEV images. Columns from left to right: (1) lidar input, (2) inpainted image, (3) stereo input, (4) inpainted image, (5) ground truth (GT). For the obstacle maps, black pixels are obstacles and white pixels are free space. \vspace{-.2in}}
  \label{fig:preds_and_obs}
\end{figure}

\subsection{Qualitative Evaluation}
Figure \ref{fig:preds_and_obs} shows qualitative results for three images of different scenes (45, 150, 285) from the test set. Results are from the input BEV images from lidar depth described in section \ref{sec:dataset} (Figure \ref{fig:preds_and_obs} columns 1, 2) and also input BEV images from stereo depth described in section \ref{sec:stereo_data} (Figure \ref{fig:preds_and_obs} columns 3, 4). Figure \ref{fig:class_distribution} shows the color key for semantics in Figure \ref{fig:preds_and_obs}. Note that the input images are based only on a single sensor measurement, while the ground truth is based on all measurements for each sequence. This is clear for image 45 in Figure \ref{fig:preds_and_obs}, where there is much less vegetation (dark green pixels) in the top left of the image compared to the ground truth, because the ground truth takes all past and future measurements into account.

For input images from lidar depth, the network reasonably fills in road pixels (pink). For image 150, the network is able to fill in the sidewalk (purple) enclosed by vegetation (green) and fence (orange) in the center of the image next to the road. For image 285, the network does not accurately predict the sidewalk perpendicular to the road in the middle of the image. However, it does a good job of filling in the bottom half of the image with vegetation and buildings (yellow). Due to the small amount of available data for training, the network does not learn all the possible structures based on the input images but reasonably fills in empty spaces occupied by road, buildings, and vegetation. 

For input images from stereo depth with semantic labels from the deeplabv3+ network, the images are generated from a different distribution than the trained inpainting network. Thus, it is expected that the network will perform worse than on the lidar derived images. In image 45 in Figure \ref{fig:preds_and_obs}, the network is unable to predict the road structure well and fills most of the unobserved space as terrain. In images 150 and 285, the road structure is cutoff early, but the network is able to predict the building structures in the bottom center of the maps. Because the stereo depth inputs are from a different distribution than the lidar depth inputs, an exact comparison cannot be made between the two as the network is trained on only ground truth lidar data. However, the potential of the inpainting network to be applied to a single stereo camera sensor is demonstrated.

\subsection{Quantitative Evaluation}
 
 Quantitative evaluation is conducted on (1) network inpainting performance compared to ground truth and (2) path planner performance using inpainted images. While planner performance improvement using inpainting is the goal of this paper, inpainting network performance is provided as a reference to the reader. 
 
 For evaluation of the inpainting performance of the network, the mean Intersection-over-Union (mIoU) is used:
 \begin{equation}
 \begin{small}
 \frac{1}{C}\sum_{c=1}^{C}\frac{\textrm{TP}_c}{\textrm{TP}_c + \textrm{FP}_c + \textrm{FN}_c}
 \end{small}
 \end{equation}

 where $\textrm{TP}_c$ is the number of true positives for each class $c$, $\textrm{FP}_c$ is the number of false positives, $\textrm{FN}_c$ is the number of false negatives, and $C$ is the number of classes. Only pixels within the inpainting mask or convex hull are used to calculate the mIoU.
 
 The mIoU for the testing dataset is 13.10. As a comparison, the top performing baseline for \textit{scene completion} from SemanticKITTI had a mIoU of 17.70. Note that the SemanticKITTI scene completion task is 3D semantic voxel completion with a dataset of 19,130 input and target pairs used for training. The mIoU measured in this paper is for image inpainting with a total of 4,408 input and target pairs available for training.  While the mIoU for inpainting is relatively low, it should greatly increase given a larger dataset.
 
 To evaluate the performance of the framework in creating better maps for path planning, experiments are conducted with an A\textsuperscript{*} planner. A custom simulator is used, where the robot is assumed to have a 360$^{\circ}$ sensor within a range of 30 pixels that can be used to update the initial lidar-based map, and replans whenever the path is discovered to be crossing through an obstacle (based on the ground truth map).
 
 The obstacle maps shown in Figure \ref{fig:preds_and_obs} are used for planning. These images were chosen to represent three levels of difficulty: easy (image 45), medium (image 150), and hard (image 285). Difficulty level is decided based on number of classes and structure of the scene. Image 285 is considered more difficult than image 150 because image 285 has cars (light blue pixels), more buildings (yellow pixels), and contains an intersection structure. For the planner, each pixel of the image is considered a node and the euclidean distance from the path node to the goal is used as the distance heuristic. Images are resized to 25\% of their original size (175$\times$145 pixels) to reduce computation time. The obstacle maps are evaluated for (1) the final number of path steps taken from start to goal, (2) the number of times the planner replans, and (3) the number of times the planner replans before determining no final path is found.

\begin{table}[]
\centering
\begin{small}
\caption{Evaluation of inpainting for planning with the baseline and ground truth (GT). Metrics are given as mean (std). Each study has 50 trials with goal positions chosen randomly. Smaller numbers are better.}
\label{tab:path_eval}
\begin{tabular}{|c|c|c|c|}
\hline
\textbf{\begin{tabular}[c]{@{}c@{}}Image 45 (easy), Fixed Start\end{tabular}}                        & \textbf{Input (baseline)} & \textbf{Inpainted (ours)} & \textbf{GT}  \\ \hline
Path Steps                                                                                                 & 70.8 (54.6)               & 75.3 (67.3)               & 64.9 (35.4)  \\ \hline
Replans                                                                                                    & 3.5 (8.1)                 & 3.2 (8.6)                 & 0 (0)        \\ \hline
Replans (no path found)                                                                                    & 53.0 (1.0)                & 43.5 (1.5)                & 0 (0)        \\ \hline
\textbf{\begin{tabular}[c]{@{}c@{}}Image 45 (easy), Random Start\end{tabular}}    & \textbf{Input}            & \textbf{Inpaint}          & \textbf{GT}  \\ \hline
Path Steps                                                                                                 & 118.6 (84.0)              & 113.5 (80.4)              & 91.1 (49.7)  \\ \hline
Replans                                                                                                    & 11.04 (12.2)              & 9.7 (10.9)                & 0 (0)        \\ \hline
Replans (no path found)                                                                                    & 48.0 (0.0)                & 42.0 (0.0)                & 0 (0)        \\ \hline
\textbf{\begin{tabular}[c]{@{}c@{}}Image 150 (medium), Fixed Start\end{tabular}}                     & \textbf{Input}            & \textbf{Inpaint}          & \textbf{GT}  \\ \hline
Path Steps                                                                                                 & 173.9 (171.3)             & 105.6 (94.1)              & 86.4 (45.5)  \\ \hline
Replans                                                                                                    & 10.4 (11.7)               & 6.1 (6.3)                 & 0 (0)        \\ \hline
Replans (no path found)                                                                                    & 5.5 (1.6)                 & 4.0 (0.9)                 & 0 (0)        \\ \hline
\textbf{\begin{tabular}[c]{@{}c@{}}Image 150 (medium), Random Start\end{tabular}} & \textbf{Input}            & \textbf{Inpaint}          & \textbf{GT}  \\ \hline
Path Steps                                                                                        & 125.8 (134.8)             & 121.7 (128.7)             & 95.5 (79.1)  \\ \hline
Replans                                                                                                    & 10.1 (15.6)               & 7.7 (10.9)                & 0 (0)        \\ \hline
Replans (no path found)                                                                                    & 24.3 (18.8)               & 19.6 (13.5)               & 0 (0)        \\ \hline
\textbf{\begin{tabular}[c]{@{}c@{}}Image 285 (hard), Fixed Start\end{tabular}}                       & \textbf{Input}            & \textbf{Inpaint}          & \textbf{GT}  \\ \hline
Path Steps                                                                                        & 149.5 (158.9)             & 95.3 (56.3)               & 91.6 (52.0)  \\ \hline
Replans                                                                                                    & 9.2 (12.5)                & 5.2 (6.2)                 & 0 (0)        \\ \hline
Replans (no path found)                                                                                    & 42.6 (8.0)                & 40.7 (10.6)               & 0 (0)        \\ \hline
\textbf{\begin{tabular}[c]{@{}c@{}}Image 285 (hard), Random Start\end{tabular}}   & \textbf{Input}            & \textbf{Inpaint}          & \textbf{GT}  \\ \hline
Path Steps                                                                                        & 169.3 (123.0)             & 140.6 (97.7)              & 124.4 (68.4) \\ \hline
Replans                                                                                                    & 15.1 (14.7)               & 7.5 (7.0)                 & 0 (0)        \\ \hline
Replans (no path found)                                                                                    & 46.1 (21.2)               & 32.4 (12.1)               & 0 (0)        \\ \hline
\end{tabular}
\end{small}
\vspace{-.2in}
\end{table}

Table \ref{tab:path_eval} displays results for the quantitative path planner evaluation. Trials were run and compared for the single lidar scan input image, the inpainted image, and the ground truth (GT). For each of the three scenes (45, 150, 285), trials were conducted using random goal locations, and start locations either random or fixed near the sensor with a Gaussian distribution. For each study, 50 trials were conducted, with the mean (std) given in Table \ref{tab:path_eval}. The inpainted image outperforms the single lidar scan for every experiment except for the easy image (45) with starting locations fixed near the sensor. This is due to the simplicity of the scene, where the inpainting image is not necessary to fill in any occluded spaces. The inpainted image outperforms the baseline in path steps taken, number of times replanned, and number of replans before discovering no path is available, for all other experiments. Inpainting allows the planner to "see" around corners and behind obstacles to predict what types of semantic labels are likely to be present in those unobserved spaces. This allows the planner to plan more informed paths which are less likely to run into dead ends. This is evident as not only do the number of path steps decrease when using the inpainted map, so do the number of times the planner needs to replan. By using previous experience (training data) to predict unknown spaces for path planning, the proposed framework imitates human behavior and greatly improves performance.

\begin{wrapfigure}{R}{0.7\textwidth}
\vspace{-.5in}
  \includegraphics[scale=0.25]{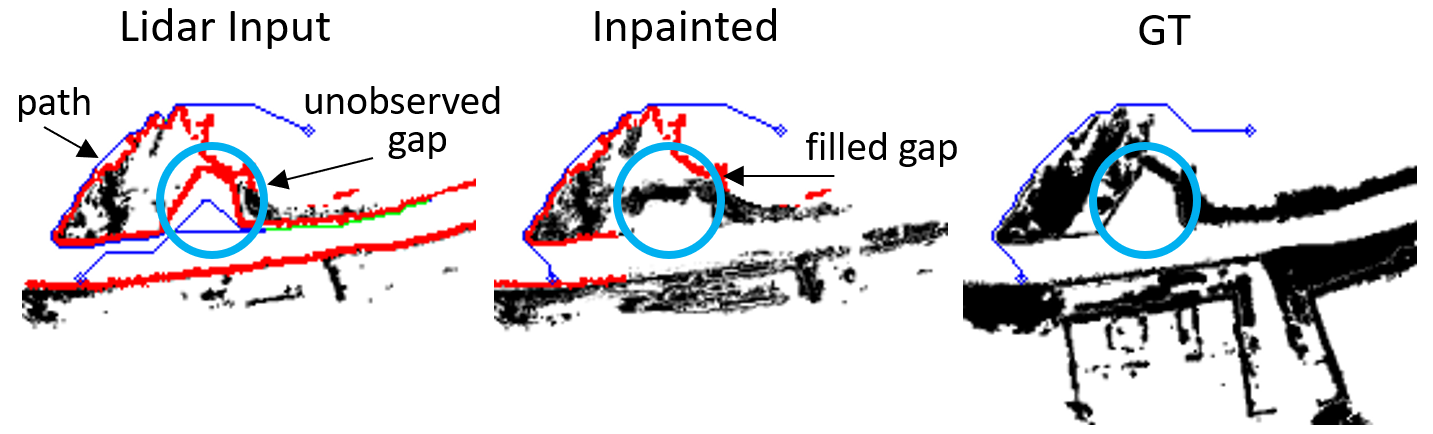}
  \caption{A comparison of a path planner trial on the lidar input obstacle map, the inpainted obstacle map, and the GT map shown in figure \ref{fig:preds_and_obs} (image 150). The path is in blue pixels. The start is in the middle left and the goal is in the top right. Red pixels are obstacles the robot observes online. Green pixels show where the path overlaps itself. The light blue circles show a gap in the lidar input map which is filled in on the inpainted image.}
  \label{fig:trial_comp_150}
\end{wrapfigure}

The ability to fill in occluded sections of vegetation or walls and buildings is incredibly helpful for path planning, as shown in Figure \ref{fig:trial_comp_150}. There, a trial on image 150 (medium difficulty) with start locations near the sensor is shown. For the lidar input obstacle map, there is initially an empty space in the middle of the map shown by the light blue circle. The planner attempts to plan a path through the gap but eventually observes there is a wall; the robot replans by continuing to move right and runs into a dead end. Eventually, the robot must plan all the way back around causing it to take many (399) additional steps. The inpainted obstacle map (Figure \ref{fig:trial_comp_150}) fills in that gap and the robot eventually takes a path (129 steps) much closer to the ground truth shortest path.


\section{Conclusion}
In this paper, a novel framework is presented to model unobserved and occluded regions for planning in outdoor urban environments. The planner uses GAN-based inpainting to generate better informed obstacle maps for planning compared to an initial sensor detection. Experiments validate the ability of the novel framework to reduce the number of steps taken, the number of times replanned, and the number of times replanned before discovering no possible path. The impact of the inpainting is higher as the scene becomes more challenging with more occlusions, as it allows the planner to predict obstacles and dead ends in unobserved spaces and avoid them.

	





\clearpage
\acknowledgments{We thank the reviewers for their helpful comments and feedback. This work is funded by the ONR under the PERISCOPE MURI Grant
N00014-17-1-2699.}

\bibliography{example}  

\end{document}